\documentclass[conference, 12pt,onecolumn,draftclsnofoot]{IEEEtran}
\pdfoutput=1
\usepackage{amsmath,amssymb,amsfonts}
\usepackage{algorithm}
\usepackage{algorithmicx}
\usepackage[noend]{algpseudocode}
\usepackage{graphicx}
\usepackage{textcomp}
\usepackage{xcolor}
\usepackage{hyperref}
\usepackage{comment}
\usepackage{mathtools}
\usepackage[caption=false, font=footnotesize]{subfig}

\newcommand{\bmf}{\mathbf f}

\usepackage{bbm}

\usepackage{longtable}
\usepackage[acronym]{glossaries}

\newacronym{dnn}{DNN}{Deep Neural Network}
\newacronym{gcn}{GCN}{Graph Convolutional Network}
\newacronym{gnn}{GNN}{Graph Neural Networks}
\newacronym{kfac}{KFAC}{Kronecker-Factored Approximate Curvature}
\newacronym{ngd}{NGD}{Natural Gradient Descent}
\newacronym{nggd}{NGGD}{Natural Graph Gradient Descent}
\newacronym{nn}{NN}{Neural Network}
\newacronym{sgd}{SGD}{Stochastic Gradient Descent}

\usepackage{amsthm}

\usepackage{biblatex}
\begin{document}

\title{Training Graph Neural Networks by
Graphon Estimation}

\makeatletter
\newcommand{\linebreakand}{%
  \end{@IEEEauthorhalign}
  \hfill\mbox{}\par
  \mbox{}\hfill\begin{@IEEEauthorhalign}
}
\makeatother

\author{
\IEEEauthorblockN{Ziqing Hu \IEEEauthorrefmark{1}} 
\IEEEauthorblockA{zhu4@nd.edu}
\and
\IEEEauthorblockN{Yihao Fang \IEEEauthorrefmark{1}} 
\IEEEauthorblockA{yfang5@nd.edu}
\and
\IEEEauthorblockN{Lizhen Lin \IEEEauthorrefmark{1}}
\IEEEauthorblockA{lizhen.lin@nd.edu}
\linebreakand
\IEEEauthorrefmark{1}Department of the Applied and Computational Mathematics and Statistics\\
\textit{University of Notre Dame} \\
Notre Dame, IN, USA
}

\maketitle
\thispagestyle{plain}
\pagestyle{plain}

\begin{abstract}

In this work, we propose to train a graph neural network via resampling from a graphon estimate obtained from the underlying network data. More specifically,  the graphon or the link probability matrix of the underlying network is first obtained from which a new network will be resampled and used during the training process at each layer.  Due to the uncertainty induced from the resampling, it helps mitigate the well-known  issue of over-smoothing in a graph neural network (GNN) model. Our framework is general, computationally efficient, and conceptually simple. Another appealing feature of our method is that it requires minimal additional tuning during the training process. Extensive numerical results show that our approach is competitive with and in many cases outperform the other over-smoothing reducing GNN training methods.  

\end{abstract}

\begin{IEEEkeywords}
Graph neural network, Graphon estimation, Oversmoothing, Resampling.
\end{IEEEkeywords}

\section{Introduction} \label{sec:intro}

This paper considers an approach for mitigating the well-known problem of over-fitting and over-smoothing in the training of  \acrfull{gnn}s and lies at the intersection of graphon estimation and neural network models. \acrshort{gnn}s, initially proposed to capture graph representations in neural networks \cite{scarselli2008graph}, have witnessed an upsurge for semi-supervised learning in a variety of tasks including node classification, link predictions, and many others.
The goal of each \acrshort{gnn} layer is to transform features while considering the graph structure by aggregating information from connected or neighboring nodes. When there is only one graph, the goal of node classification  is to predict node labels in a graph while only a portion of node labels are available (even though the model might have access to the features of all nodes). Inspired by the advance of convolutional neural networks \cite{lecun1998gradient} in computer vision \cite{krizhevsky2012imagenet}, \acrfull{gcn} \cite{kipf2016semi}  employs the spectra of graph Laplacian for filtering signals and the kernel can be approximated using Chebyshev  polynomials or functions \cite{zhou2018graph, wu2020comprehensive}.  GCN has become a standard and popular tool in the emerging field of geometric deep learning \cite{bronstein2017geometric}. However, the issue of over-fitting arises 
when an overparametrized  model  such as the deep neural network, is applied to a distribution with limited training data, where the learned fits the training data well but generalizes poorly to the testing data. This can be illustrated briefly by fitting a deep \acrshort{gnn} (more than 4 layers) to small a graph data (e.g., the Cora dataset). On the other hand, the issue of over-smoothing introduced by \cite{li2018deeper}  towards the other extreme, bringing difficulties to deep \acrshort{gnn} training. Further explained by \cite{wu2020comprehensive}, graph convolutions mix representations of adjacent nodes and result in all nodes' representations converging to a stationary subspace or point \cite{oono2019asymptotic}. This phenomenon is called \emph{over-smoothing} of node features \cite{rong2019dropedge}. By way of illustration, GCN models with more than 8 layers are observed to converge poorly in our experiments.

To alleviate those two issues, inspired by \cite{zhang2017estimating}, we propose a new  \acrshort{gnn}   structure with resampling the adjacency matrix in the feed forward propagation via graphon estimation. Graphon, a function that determines the matrix of edge probabilities, plays an important  role in graph theory and statistics \cite{optimal-graphon, zhao2019change}. The estimation of probabilities of network edges from the observed adjacent matrix, known as ''graphon estimation'', has a wide range of applications to predicting missing links and network denoising \cite{Chatterjee_2015,10.1214/15-AOS1370}. In our framework, we assume the observed adjacency matrix $A$ is generated from an underlying probability matrix $P$ so that for $i\leq j$, $A_{ij}'s$ are independent Bernoulli($P_{ij}$) trails where $P_{ij}$ are edge probabilities. Consequently, we resample the adjacency matrix $A$ from the estimated distribution $P$ in the feed forward propagation for each training epoch. There are several benefits in applying the resampling strategy for training \acrshort{gnn}. First, resampling the adjacency matrix is one way for data augmentation to relieve the over-fitting. We obtain more graph samples from the underlying distribution under this method. Second, resampling strategy can be considered as noise addition to the deterministic \acrshort{gnn} and which avoids our nodes' representations converging to the stationary subspace  \cite{oono2019asymptotic}, hence solving the over-smoothing phenomena. Finally, since we consider the underlying distribution of the graph, our method is able to achieve a stable result under noisy graphs.         

Our work is organized as follows. Section \ref{sec-related} reviews some related work. Section \ref{sec:bg} provides an overview of some background information such as \acrshort{gnn} and GCN. The proposed algorithm is described in section \ref{sec:method} and a series of experiments are performed in section \ref{sec:exp} to evaluate our proposed method's efficiency and sensitivity to hyper-parameters.
Finally, the work is concluded in section \ref{sec:conc}.

\section{RELATED WORKS}
\label{sec-related}
\subsection{GRAPH NEURAL NETWORK}
Most graph neural networks, as mentioned above, are treating the related graph as ground-truth deterministic structure between nodes, but often the graph itself may be subjected to random perturbation or theoretical assumptions that might lead to  unreliable results given the uncertain graph. \cite{zhang2019bayesian} firstly propose a Bayesian version GCN (BGCN) to incorporate the potential uncertainty presented in the graph. Similarly, \cite{elinas2019variational} extend the BGCN to include the node features and adopt the variational inference method to estimate the posterior distribution which achieve comparable result under  adversarial attack setting. However, due to the computation complexity, it's not easy to apply the model on large datasets. Based on bilevel programming, \cite{franceschi2019learning} proposes a method for jointly learning the graph structure and network parameter via constrained optimization. From over-smoothing alleviation perspective, \cite{hasanzadeh2020bayesian} propose Graph DropConnect (GDC) method to alleviate the over-smoothing issue in GCN by resampling the graph for each node feature and show that DropOut \cite{srivastava2014dropout}, DropEdge \cite{rong2019dropedge} and Node Sampling \cite{chen2018fastgcn} are special cases of GDC with respect to different settings. However, there is no theoretical guarantee that GDC can reduce the over-smoothing issue. Finally, similar to our work, \cite{zhao2020data} propose a two-step procedure for data augmentation in graph neural network. They firstly use graph auto-encoder (GAE) \cite{kipf2016variational} to estimate the edge probability which is used for resampling in later procedure. Then, combining  the resampled graph with original graph, they applied another graph neural network to learning the embedding of nodes. However, their emphasis is very different from ours as we focus on reducing over-smoothing issue in a deep graph neural network.

\subsection{GRAPHON ESTIMATION}

Graphon estimation is an important component of our proposed procedure for training the GNNs. A prominent estimator of the graphon is the so-called USVT (Universal Singular Value Thresholding) estimator \cite{Chatterjee_2015}. USVT is a general procedure for estimating the entries of a large structured matrix, given a noisy realization of the matrix. This includes estimating the link probability matrices  which is our case of interest.  The key idea behind USVT is to threshold the singular values of the observed matrix at an universal threshold which essentially approximates the rank of the population matrix, and then compute an approximation of the population matrix using the top singular values and vectors. 
A recent work by \cite{zhang2017estimating} proposes a  statistically consistent and computationally efficient method  for estimating the link probability matrix by neighborhood smoothing. 
More specifically, given an adjacent matrix $A$, the link probability $P_{ij}$ between node $i$ and $j$ is estimated by 
\begin{align}
\label{eq:NBS}
\hat{P}_{ij}=\frac{1}{2}\left(\frac{\sum_{i'\in \mathcal N(v_i)} A_{i'j}}{|\mathcal N(v_i)|}+\frac{\sum_{j'\in \mathcal N(v_j)} A_{ij'}}{|\mathcal N(v_j)|}\right),
\end{align}
where $\mathcal N(v_i)$ is a certain set of neighboring nodes of node $v_i$ (which consists of the nodes that have similar connection patterns as node $v_i$).  Rather than simply choosing connected node as neighbours, the neighbour is selected by the following criteria $\mathcal{N}(v_i)=\left\{i^{\prime} \neq i: \tilde{d}\left(i, i^{\prime}\right) \leq q_{i}(h)\right\}$ where distance $\tilde{d}^{2}\left(i, i^{\prime}\right)$ is defined as $\tilde{d}^{2}\left(i, i^{\prime}\right)=\max _{k \neq i, i^{\prime}}\left|\left\langle A_{i}-A_{i^{\prime}}, A_{k \cdot}\right\rangle\right| / n$ and $q_{i}(h)$ is the $h$-th sample quantile of the set $\left\{\tilde{d}\left(i, i^{\prime}\right): i^{\prime} \neq i\right\}$.

Typically for large networks USVT is more scalable than the neighborhood-smoothing approach. There are several other methods for graphon estimations, e.g., by fitting a stochastic blockmodel \cite{wolfe2013nonparametric}. These methods can also be used in our proposed GNN training algorithm.   


\section{Notation and Background} \label{sec:bg}
\subsection{Notation}
Let $\mathcal{G}=(\mathcal{V}, \mathcal{E})$ represent the input graph with node set $\mathcal{V}$ of size $N$ and edge set $\mathcal{E}$ where $v_i \in \mathcal{V}$ and $(v_i, v_j) \in \mathcal{E}.$ $N(v_i)$ denotes all the neighbours connected to node $v_i$. We denote the $\boldsymbol{X}=\left\{\boldsymbol{x}_{1}, \cdots, \boldsymbol{x}_{N}\right\} \in \mathbb{R}^{N \times f}$ as the node feature matrix and $\mathbf{A} \in\{0,1\}^{N \times N}$ as the adjacency matrix. Let $D$ be the diagonal matrix with node degrees $d_i = \sum_{i = 1}^{N(v_i)}A_{i,j}$ as its entries. $I$ is the identity matrix.

\subsection{Graph Neural Networks}
The \acrfull{gnn} can be seen as an extension of \acrshort{nn} that learns the embedding of the data in graph domains \cite{scarselli2008graph}.
The basic idea can be written by a local transition function as, for each node $v_i, \dots, v_n$, 
\begin{equation} \label{eq:layer2}
    \boldsymbol{x}_{v_i}^{l} = \bmf_{l}(\boldsymbol{x}_{v_i}^{l - 1}, \underline{\boldsymbol{x}}_{N(v_i)}^{l - 1}; W_l)
\end{equation}
where $\underline{\boldsymbol{x}}_{N(v_i)}^{l - 1}$ represents all the neighbouring information of node $v_i$ at the $l$th layer. The $\boldsymbol{x}_{v_i}^{l}$ and $ W_l$ are the embedding of node $v_i$ and the  model parameters at $l$-th layer, respectively.

The \acrfull{gcn} developed in \cite{kipf2016semi} is one of the variants of \acrshort{gnn} with the message passing mechanism as the graph signal filter in graph Fourier space, which can be written in matrix form as:
\begin{equation}
    \boldsymbol{X}^{l} = \sigma\left(\tilde{A} \boldsymbol{X}^{l-1}W^{l}\right),
\end{equation}
where $\sigma$ is a element-wise nonlinear activation function such as $\text{ReLU}(x) = \text{max}(x, 0)$, $W^{l}$ is a $f^{l}\times f^{l-1}$ parameter matrix that needs to be estimated.
$\tilde{A}$ denotes the normalized adjacency matrix defined by $\tilde{A} = (D+I)^{-1/2}(A+I)(D+I)^{-1/2}$.



\section{Method} \label{sec:method}
In this section, we introduce the methodology of graphon estimation in training of generic GNNs. Moreover, we also propose and  implement its layer-wise variant where we resample the adjacent matrix $A$ from the estimated distribution $P$ for each layer in the model. We also illustrate how our graphon estimation technique can alleviate over-smoothing and over-fitting issues. 

\subsection{Resampling strategy}

For the given graph $\mathcal{G}=(\mathcal{V}, \mathcal{E})$ with adjacency matrix $A$, we apply the neighbouring smoothing method (NBS) \cite{zhang2017estimating} that was described in Section \ref{sec-related} to estimate the underlying link probability matrix $P$, denoted as $\hat{P}$ (see equation \eqref{eq:NBS}). Other graphon estimation methods can also be used.  At each training epoch, we resample a new adjacency matrix $\hat{A}$ from the estimated link probability matrix $
\hat{P}$ element-wisely following Bernoulli distribution:
\begin{equation} \label{eq:resample}
    \hat{A}_{ij} \sim Bern(\hat{P}_{ij}), \quad 1\leq i,j \leq n.
\end{equation}
We replace $A$ with $\hat{A}$ in equation \eqref{eq:resample} during training. The original $A$ is utilized for validation and test.

    

    

\subsection{Layer-wise variant}
Besides resampling the adjacency matrix $\hat{A}$ for the whole propagation, we can resample $\hat{A}^l$ independently from Equation \ref{eq:resample} for each $l$-th layer. In particular, different $l$-layer could have different matrix $\hat{A}^l$ and additional randomness and augmentation of the original data could be brought to our training process. We compare its performance with the vanilla resampling strategy in Section \ref{sec:exp}.

\subsection{Alleviating over-smoothing and over-fitting}
Over-fitting occurs when an overparametrized model is utilized to fit a distribution with limited training data. To prevent this issue, we first estimate the underlying graphon of the input graph. Our resample strategy works as a data augmentation technique by generating different realizations of the input data from the underlying distribution. On the other hand, the over-smoothing phenomenon indicates that the node features would converge to the fixed point as the network depth increases \cite{li2018deeper}. Furthermore, \cite{oono2019asymptotic} has extended the original explanation to a more general framework by considering the non-linear activation function in the GCN propagation. Instead of converging to the fixed point, the node features will converge to a subspace related to the eigenspace of the graph adjacency matrix. The key point of the theory illustrates that when the same adjacency matrix is utilized for all layers, the whole dynamic system will go closer to the corresponding eigenspace as the number of layers increases under specific assumptions. To avoid the phenomena, our proposed method draws random adjacency matrices from the underlying graphon in training, which helps the dynamic system escape from the subspace. Different from other random sampling methods like DropEdge and DropNode, the graphon estimation method is able to detect the underlying graphon and provide a robust estimator with statistical bounds. Consequently, our proposed method enables us to train deep GNNs more effectively, notably when the input graphs are noisy.

\section{Experiments} \label{sec:exp}
In this section, we evaluate the proposed resampling algorithm on several datasets through different network architectures. A summary of datasets and their splitting settings are provided. All the experiments are conducted by Pytorch \cite{paszke2019pytorch} and Pytorch Geometric \cite{fey2019fast}.

\subsection{Datasets}
The summary statistics of the data are shown in Table~\ref{tab:datasets}. We follow three different data-splitting settings for semi-supervised tasks on these datasets. The first setting comes from \cite{yang2016revisiting}, named `public', in which $20$ samples for each cluster are randomly drawn for training, $500$ for validation, and $1000$ for the test. For the next split in \cite{chen2018fastgcn}, named `complete', $1708$ samples are selected for training, $500$ for validation and $500$ for test. The last setting comes from \cite{levie2018cayleynets}, named `full', which chooses all of the samples for training except for $500$ nodes for the validation and $500$ nodes for the test. For graphon estimation, we use the whole dataset and pre-compute it before running the network model.

\begin{table}[htbp]
    \centering
    \caption{Citation network datasets summary}
    \begin{tabular}{lcccc}
        Dataset & Nodes & Edges & Classes & Features\\
        \hline
        Citeseer & 3,327 & 4732 & 6 & 3,703\\
        Cora & 2,708 & 5,429 & 7 & 1,433\\
        Pubmed & 19,717 & 44,338 & 3 & 500
    \end{tabular}
    \label{tab:datasets}
\end{table}

\subsection{Architectures}
We employ 3 different widely used GNN architectures in our experiment: GCN\cite{kipf2016semi}, GraphSAGE\cite{hamilton2017inductive} and JK-NET\cite{xu2018representation} with layers ranging from $2$ to $16$. Note that, for JK-NET, the number of layers doesn't include the concatenation and output layer.  For the hidden layer dimension, we follow the same $64$-dimension setting with \cite{kipf2016semi}. We choose ReLU function as our activation function between each layer and the cross entropy as our loss function.

\subsection{Optimization}
We initialize the weight parameters through Xavier uniform initialization. All of the data are row-wise normalized accordingly \cite{glorot2010understanding}. The model is trained for $1000$ epochs with a learning rate start from $0.001$ and decreased at epoch, $300$ and $600$ with decay rate $0.5$. The Adam optimizer is used without any penalty term.

\subsection{Results}
Due to limited space, we only attach the result of `Public' splitting setting, which is given in Table~\ref{tab:public}. We let the `Resampling' represents our original algorithm while the `Layerwise' represents the layer-wise variant of our method. We pick the best result of `Dropedge' in each setting where the dropping rate ranging from $0.2$ to $0.8$. The reported value are the average and stand deviation over $10$ runs in Table \ref{tab:public}. Also, we apply the early stopping to keep track of the validation loss, if the loss stops decreasing for several epochs. As shown by the numerical results in the table, our Resampling method or its layerwise  variant performs the best in most of the settings for all three datasets and three different GNN architectures considered. Precisely, we consider a $8$-layer GCN with/without Resampling (Layerwise) on the Citation dateset. In term of the loss evolution among different methods, our Resampling method or the layerwise variant is able to alleviate both overfitting and oversmoothing issues as shown in Figure~\ref{fig:Loss}.    Similar patterns are observed in other splitting settings.  The results demonstrate the effectiveness of our  proposed methods in comparing with other state-of-art methods. In comparing with other methods, like Dropedge, which requires multiple comparisons to determine the appropriate Dropedge rate, our approach requires minimal additional tuning. Once the estimation of graphon is completed, we can re-use it without any further modification. 
\begin{figure*}[htbp]
    \centering
    \subfloat[Cora - Training\label{fig:Cora-training}]{
    \includegraphics[width=.35\textwidth]{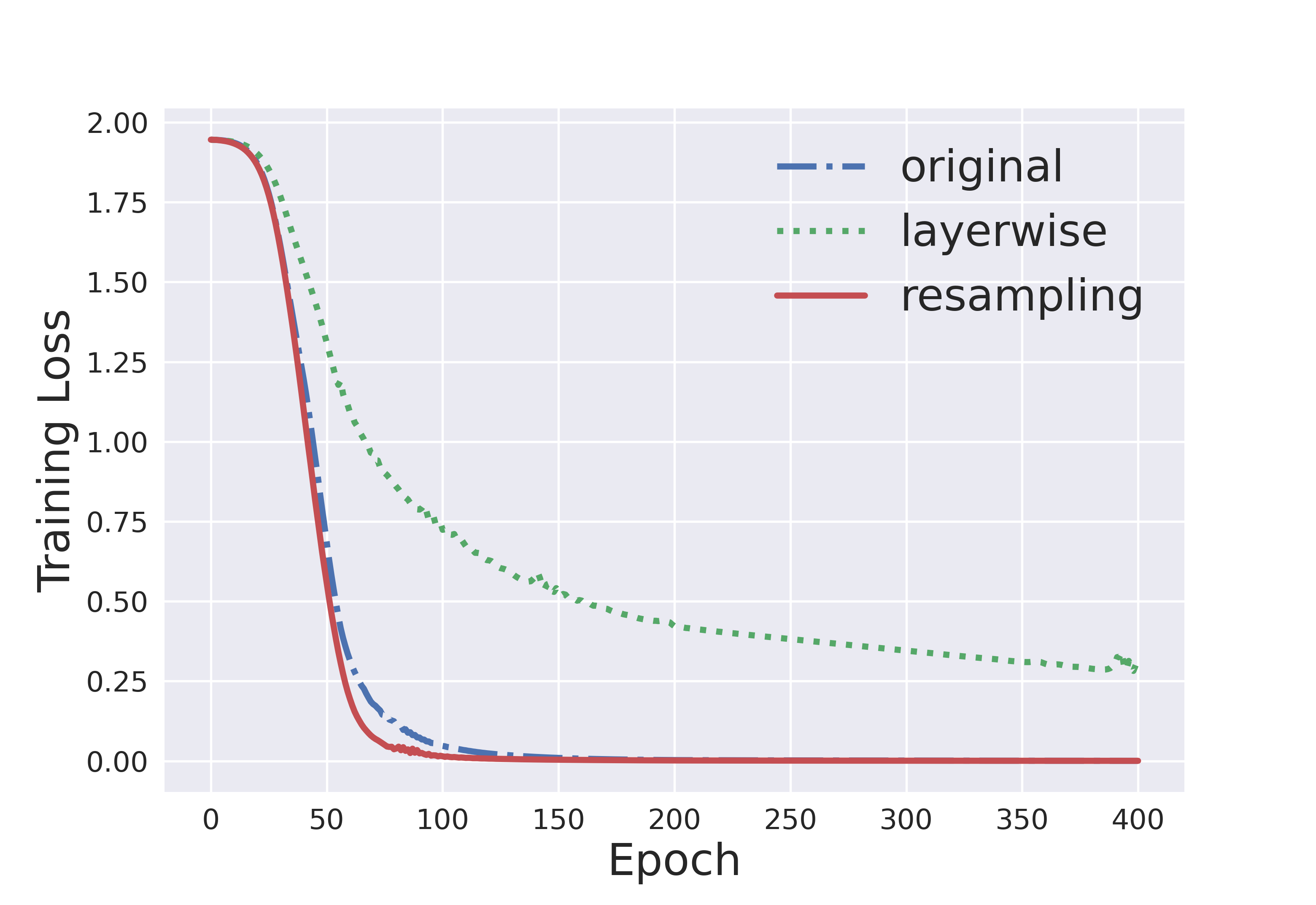}}
    \subfloat[CiteSeer - Training\label{fig:Cite-training}]{
    \includegraphics[width=.35\textwidth]{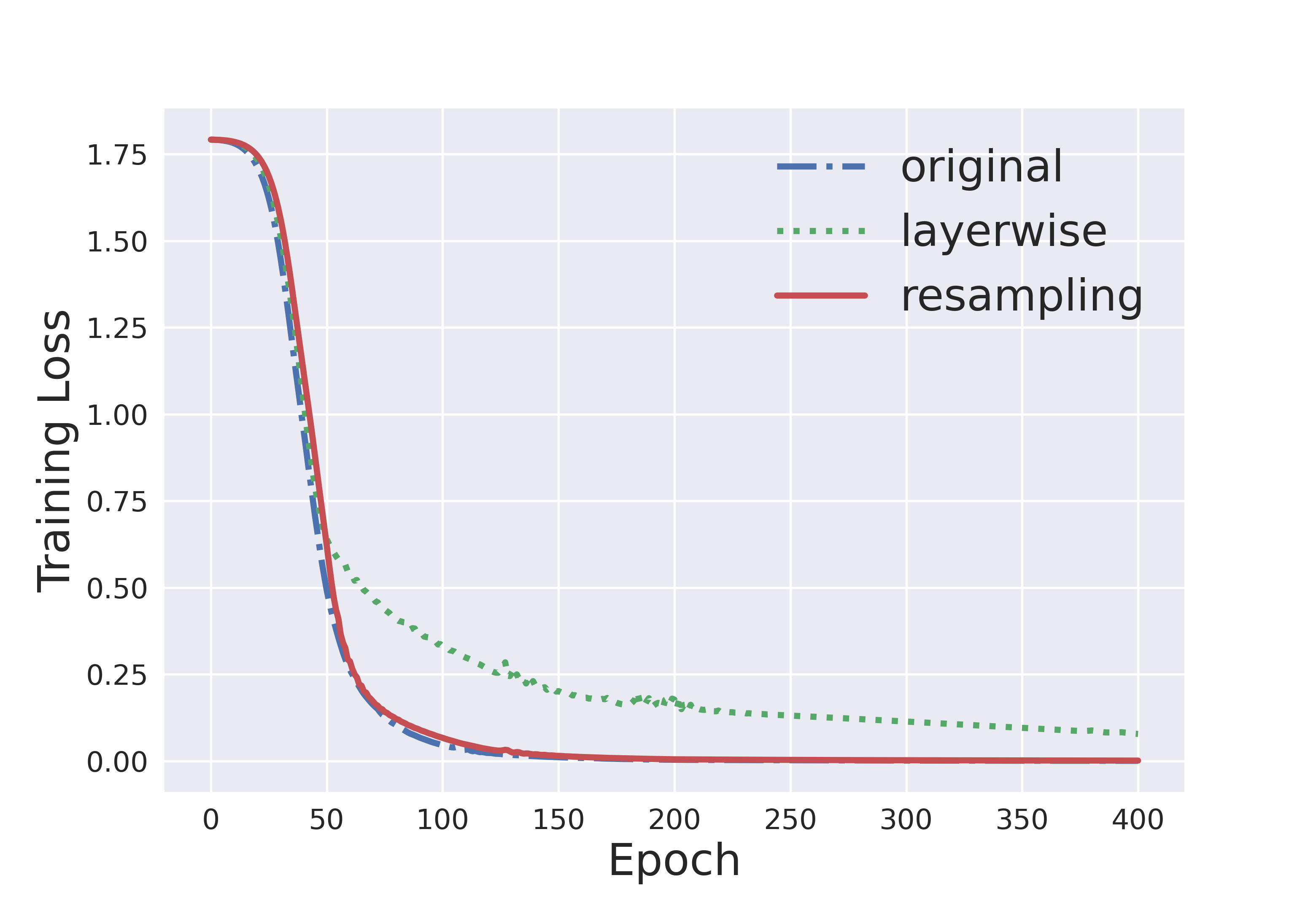}}
    \subfloat[PubMeb - Training\label{fig:Pub-training}]{
    \includegraphics[width=.35\textwidth]{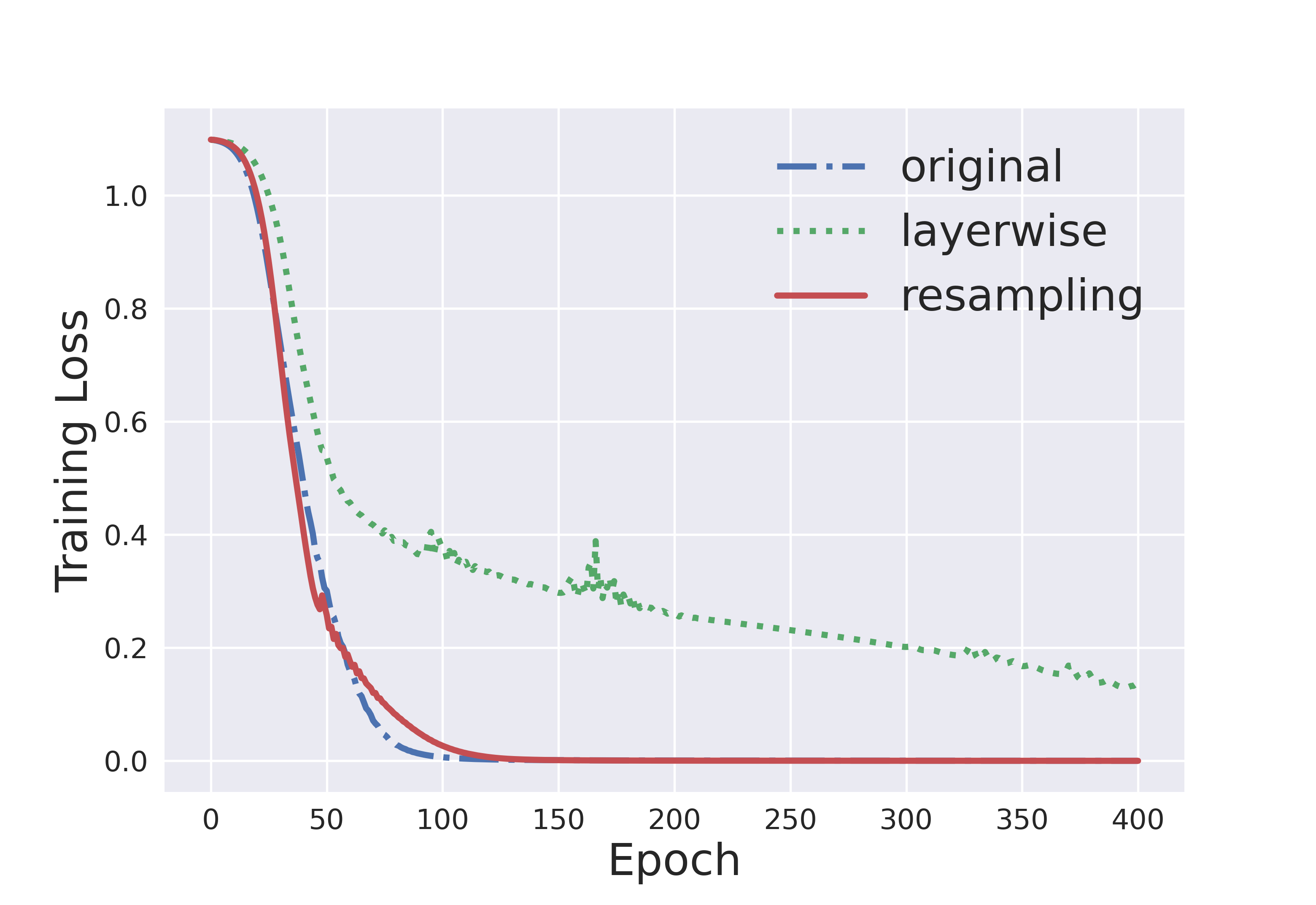}}\\
    \subfloat[Cora - Validation\label{fig:Cora-val}]{%
    \includegraphics[width=.35\textwidth]{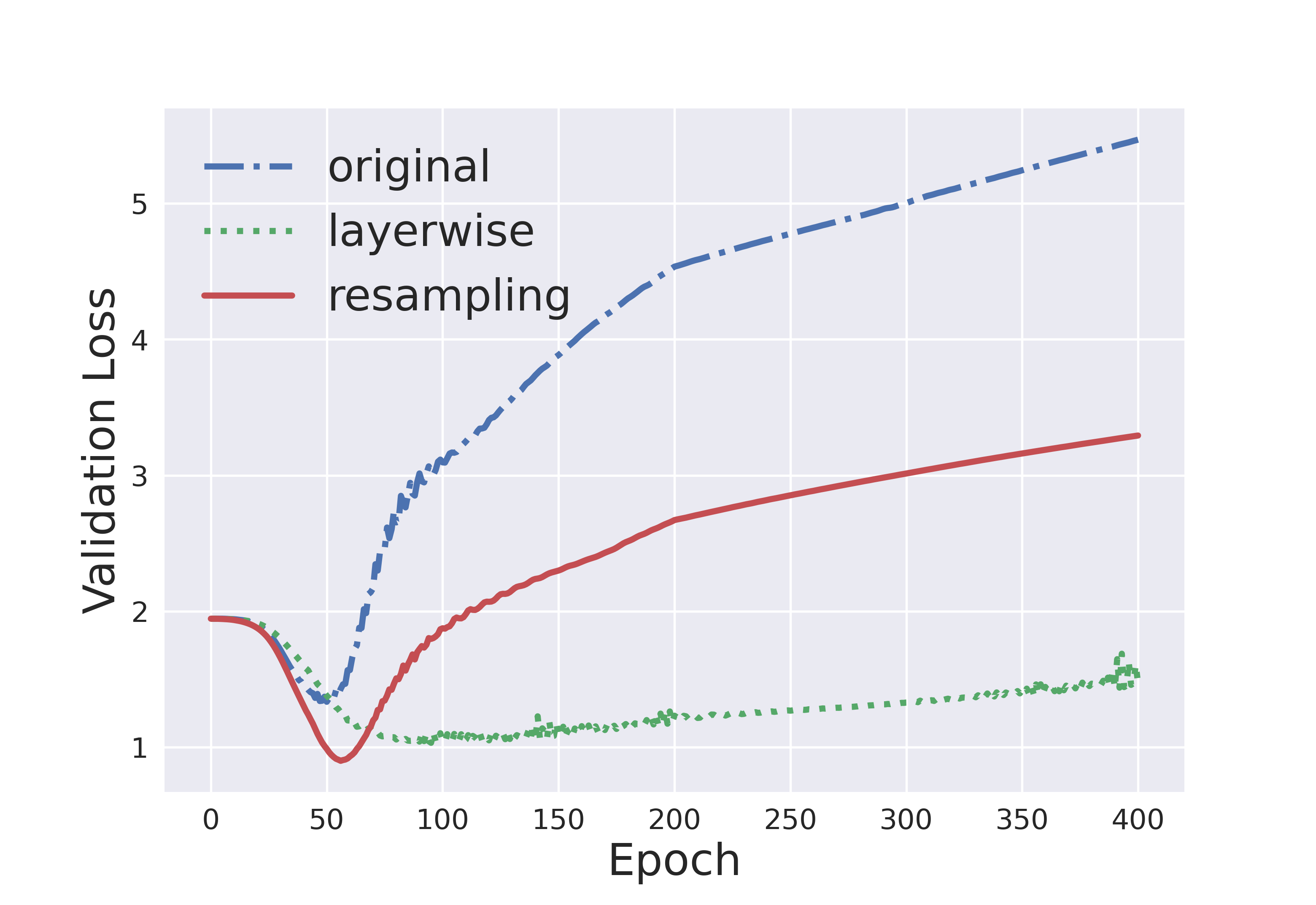}}
    \subfloat[CiteSeer - Validation\label{fig:Cite-valid}]{%
    \includegraphics[width=.35\textwidth]{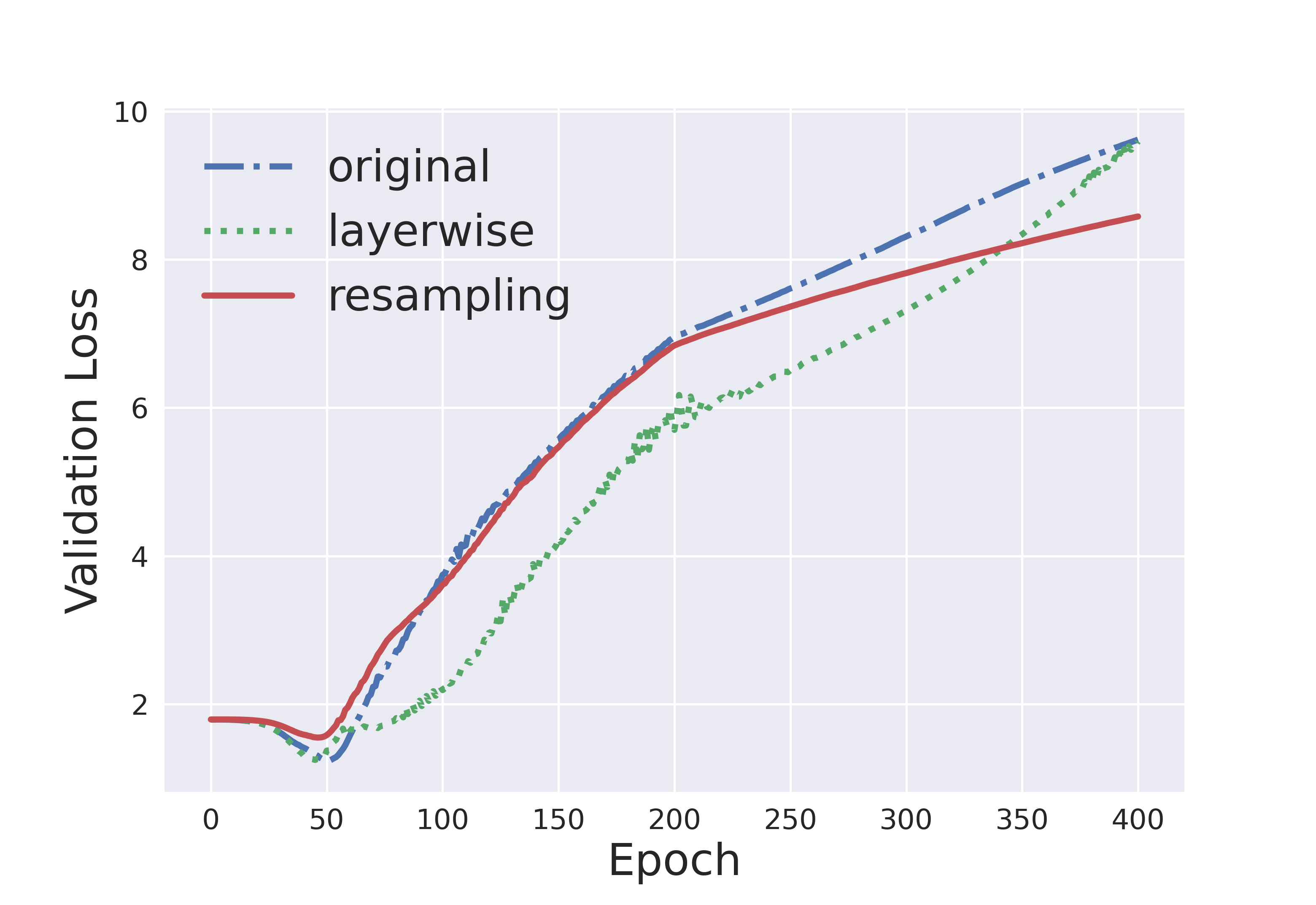}}
    \subfloat[PubMed - Validation\label{fig:Pub-valid}]{%
    \includegraphics[width=.35\textwidth]{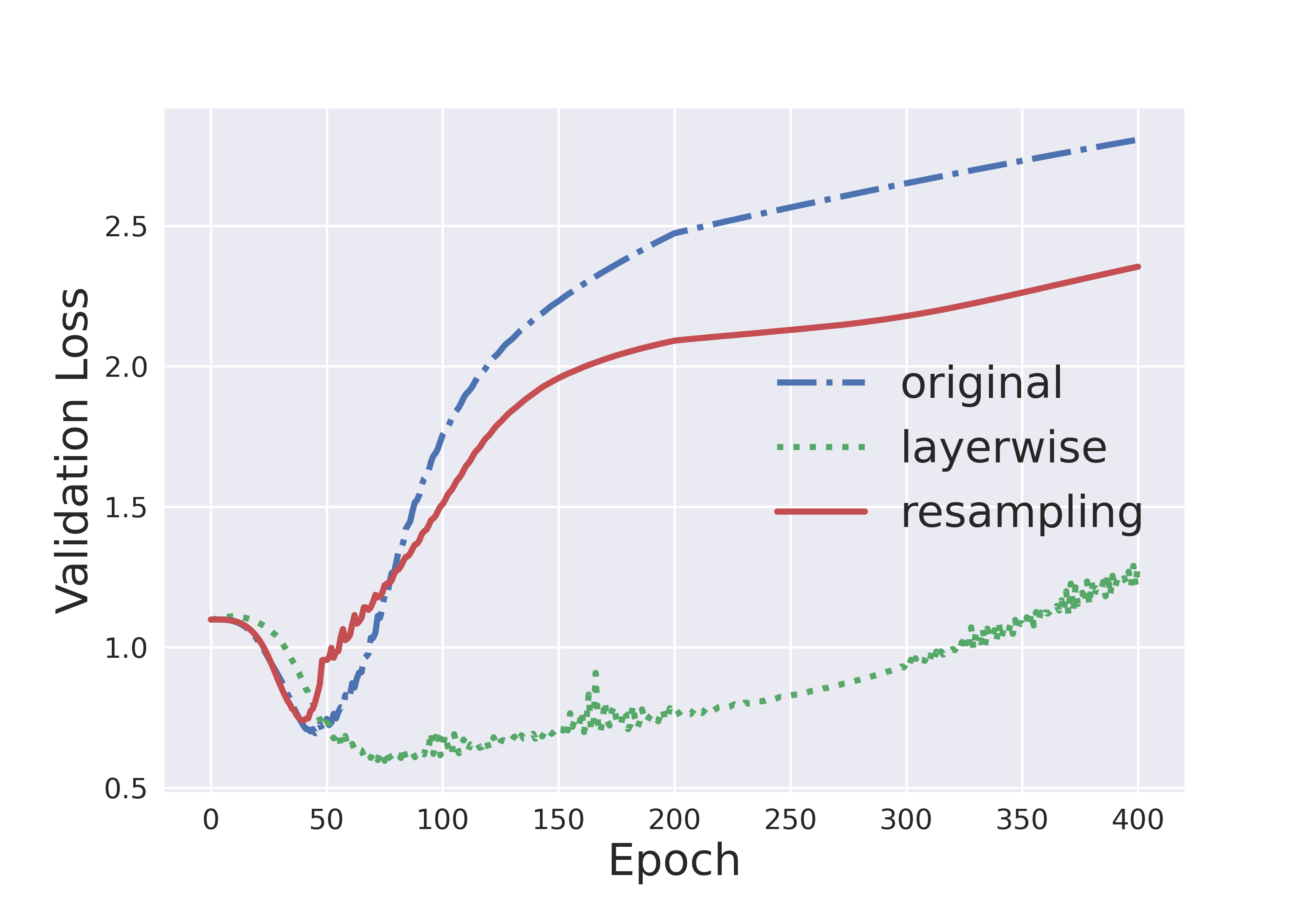}}
    \caption{The training and validation loss of GCNs on the public split of Citation datasets. We implement the original $8$-layer GCN (in blue), with resampling (in red) and with layerwise variant (in green). The original $8$-layer GCN comes up with the overfitting issue in several epochs with low training loss but high validation loss on all datasets. Furthermore, the validation of the  original $8$-layer GCN diverges significantly in all cases due to over-smoothing issue. In contrast, our Resampling and layerwise variant (in green), alleviates both overfitting and oversmoothing issues and achieves smaller validation errors in notably for Cora and PubMed Fig.~\ref{fig:Cora-val} and ~\ref{fig:Pub-valid}.
   We utilize early stop technique in training around $100$ epochs to achieve the best performance in Table \ref{tab:public}. Here we show the loss within $400$ epochs for a complete comparison.   }
  \label{fig:Loss}
\end{figure*}

\begin{table}[htbp]
    \centering
    \caption{The test accuracy  of different methods in `public' setting}
    \bgroup
    \def\arraystretch{0.8}
    \begin{tabular}{cccccc}
        \hline
        \multicolumn{6}{c}{\textbf{\textit{Cora}}}\\
        GCN & Original & Resampling & Layerwise & Dropedge & Dropout $0.2$ \\ \hline
        $2\text{-Layers}$ & $79.74 \pm 0.006$ & $ \mathbf{80.30\pm 0.004} $ &$ 76.53 \pm 0.015$ & $ 80.16 \pm 0.005$ & $ 80.56 \pm 0.006$ \\
        
        $4\text{-Layers}$ & $ 76.91 \pm  0.020 $ & $ 77.15 \pm  0.016 $ & $ 74.46\pm 0.013 $ & $ 78.93 \pm 0.014 $ &$  \mathbf{78.95 \pm  0.012} $ \\
        
        $8\text{-Layers}$ &$  68.76 \pm 0.060 $ & $ 70.93 \pm 0.040$ & $ 69.16 \pm 0.020 $ &$ 71.38 \pm 0.031 $ & $ \mathbf{71.52 \pm 0.028}$ \\ 
        
        $16\text{-Layers}$ & $ 59.36 \pm 0.039$ & $ 59.43 \pm 0.025 $ & $ \mathbf{65.31 \pm 0.037}$ & $ 60.50  \pm 0.027$ & $ 64.48  \pm 0.026 $\\ \hline
        	
        GraphSage  & Original & Resampling & Layerwise & Dropedge & Dropout $0.2$ \\ \hline
        $2\text{-Layers}$ & $76.80  \pm 0.007$ & $ \mathbf{78.16 \pm 0.006}$ & $ 78.08  \pm 0.005$ & $ 52.06 \pm 0.034$ & $ 77.05 \pm 0.010$ \\  
        $4\text{-Layers}$ & $ 78.41 \pm 0.012$ &  $78.51 \pm 0.013$ &$  77.01 \pm 0.014$ & $ 43.21 \pm 0.067$ & $ \mathbf{79.34 \pm 0.010}$ \\ 
        $8\text{-Layers}$ & $ 75.79  \pm 0.014$ & $ 74.45 \pm 0.021$ &$  74.58  \pm 0.025$ & $ 52.62 \pm 0.075$ & $ \mathbf{76.90  \pm 0.012}$ \\  
        $16\text{-Layers}$ & $ \mathbf{72.61 \pm 0.020}$ &$  71.63 \pm 0.016$ & $ 69.84 \pm 0.028$ & $ 34.21 \pm 0.095$ & $ 67.18  \pm 0.022$ \\ \hline
        JK-Net  & Original & Resampling & Layerwise & Dropedge & Dropout $0.2$ \\ \hline
        $2\text{-Layers}$ & $ 47.83 \pm 0.032$ & $ 65.32 \pm 0.007$ &$  \mathbf{65.89  \pm 0.023}$ & $ 52.52 \pm 0.041$ & $ 48.34 \pm 0.021$ \\ 
        $4\text{-Layers}$ & $ 49.58  \pm 0.037$ & $ \mathbf{63.56 \pm 0.019}$ & $ 63.49  \pm 0.016 $& $ 44.66  \pm 0.060$ & $ 51.37  \pm 0.025$ \\  
        $8\text{-Layers}$ & $ 49.77  \pm 0.017$ & $ \mathbf{62.89  \pm 0.025}$ & $ 60.88  \pm 0.016$ & $ 32.31 \pm 0.077 $& $ 55.94 \pm 0.029$ \\  
        $16\text{-Layers}$ & $ 56.72 \pm 0.027$ & $ \mathbf{59.02 \pm 0.046}$ & $ 52.89  \pm 0.027$ &$  27.46  \pm 0.075$ & $ 57.46  \pm 0.025 $ \\
        \hline     
        \multicolumn{6}{c}{\textbf{\textit{CiteSeer}}}\\
        GCN & Original & Resampling & Layerwise & Dropedge & Dropout 0.2 \\ \hline
        $2\text{-Layers} $ & $ 68.12 \pm 0.008 $ & $ 68.40 \pm 0.006 $ & $ 67.66 \pm 0.010 $ & $ \mathbf{68.69  \pm 0.005} $ & $ 68.62 \pm 0.009$ \\  
        $4\text{-Layers} $ & $ \mathbf{65.61 \pm 0.024} $ & $ 64.92 \pm 0.018 $ & $ 65.26 \pm 0.016 $ & $ 65.48  \pm 0.015 $ & $ 65.46 \pm 0.016 $ \\  
        $8\text{-Layers} $ & $ 55.29  \pm 0.029 $ & $ 52.36 \pm 0.041 $ & $ \mathbf{60.63 \pm 0.030} $ & $ 55.89 \pm 0.037 $ & $ 55.83 \pm 0.043 $ \\  
        $16\text{-Layers} $ & $ 48.06  \pm 0.027 $ & $ 47.46  \pm 0.025 $ & $ \mathbf{51.39  \pm 0.015} $ & $ 48.39  \pm 0.030 $ & $ 51.41 \pm 0.024$ \\ \hline
        GraphSage  &   Original   &   Resampling & Layerwise & Dropedge & Dropout 0.2 \\ \hline
        $2\text{-Layers} $ & $ 65.21 \pm 0.013 $ & $ 67.74 \pm 0.012 $ & $ \mathbf{68.20  \pm 0.012} $ & $ 50.64 \pm 0.037 $ & $ 66.19  \pm 0.007$ \\ 
        $4\text{-Layers}$ & $ 64.80  \pm 0.017 $ & $ 67.01 \pm 0.017 $ & $ \mathbf{67.28  \pm 0.012} $ & $ 38.25 \pm 0.085 $ &  $ 64.96 \pm 0.017$ \\  
        $8\text{-Layers}$ & $ 57.55 \pm 0.048 $ & $ 62.06 \pm 0.026 $ & $ \mathbf{63.29 \pm 0.023 }$ & $ 44.23 \pm 0.037 $ & $ 58.53 \pm 0.043$ \\  
        $16\text{-Layers}$ & $ 50.08  \pm 0.053 $ & $ 56.56 \pm 0.032 $ & $ \mathbf{56.94 \pm 0.028 }$ & $ 30.59  \pm 0.043 $ & $ 44.30 \pm 0.031$ \\ \hline
        JK-Net  &  Original  &  Resampling  &  Layerwise  &  Dropedge  &  Dropout 0.2 \\ \hline
        $2\text{-Layers}$ & $ 38.94 \pm 0.019 $ & $ 51.71 \pm 0.018 $ & $ \mathbf{52.83 \pm 0.026} $ & $ 45.40 \pm 0.032 $ & $ 40.04 \pm 0.023$ \\  
        $4\text{-Layers}$ & $ 39.83 \pm 0.020 $ & $ 50.86 \pm 0.022 $ & $ \mathbf{51.16  \pm 0.023} $ & $ 38.51 \pm 0.049 $ & $ 41.74 \pm 0.021$ \\  
        $8\text{-Layers}$ & $ 37.76  \pm 0.027 $ & $ \mathbf{48.91 \pm 0.023} $ & $ 48.26  \pm 0.015 $ & $ 27.53 \pm 0.046 $ & $ 41.59  \pm 0.029$ \\  
        $16\text{-Layers}$ & $ 43.01 \pm 0.027 $ & $ 44.04 \pm 0.032 $ & $ 43.02 \pm 0.027 $ & $ 25.55 \pm 0.035 $ & $ \mathbf{44.35 \pm 0.015}$ \\ 
        \hline
        \multicolumn{6}{c}{\textbf{\textit{PubMed}}}\\
        GCN & Original & Resampling & Layerwise & Dropedge  & Dropout 0.2 \\ \hline
        $2\text{-Layers}$ & $ 76.37  \pm 0.003 $ & $ 76.45 \pm 0.003 $ & $ 74.88  \pm 0.019 $ & $ 76.59  \pm 0.003 $ & $ \mathbf{76.69  \pm 0.002}$ \\ 
        $4\text{-Layers}$ & $ 76.75 \pm 0.004 $ & $ 76.76  \pm 0.006 $ & $ 75.75 \pm 0.011 $ & $ \mathbf{77.03 \pm 0.007} $ & $ 76.94 \pm 0.006 $\\
        $8\text{-Layers}$ & $ 74.26 \pm 0.020 $ & $ 73.78  \pm 0.027 $ & $ \mathbf{76.41 \pm 0.011} $ & $ 75.11 \pm 0.033 $ & $ 75.28  \pm 0.026 $\\ 
        $16\text{-Layers}$ & $ 71.86 \pm 0.022 $ & $ \mathbf{73.70  \pm 0.021} $ & $ 72.34 \pm 0.021 $ & $ 73.15 \pm 0.017 $ & $ 73.07  \pm 0.021 $\\ \hline
        GraphSage & Original & Resampling & Layerwise & Dropedge  & Dropout 0.2 \\ \hline
        $2\text{-Layers}$ & $ 74.93 \pm 0.006 $ & $ \mathbf{77.34 \pm 0.004} $ & $ 76.62 \pm 0.007 $ & $ 69.21 \pm 0.023 $ & $ 75.24 \pm 0.006 $\\ 
        $4\text{-Layers}$ & $ 74.60  \pm 0.009 $ & $ \mathbf{76.38  \pm 0.010} $ & $ 74.67  \pm 0.006 $ & $ 64.24 \pm 0.046 $ & $ 75.56  \pm 0.010 $\\ 
        $8\text{-Layers}$ & $ 74.72 \pm 0.013 $ & $ 74.17 \pm 0.015 $ & $ 74.06 \pm 0.015 $ & $ 69.36  \pm 0.029 $ & $ \mathbf{76.24 \pm 0.012 }$\\ 
        $16\text{-Layers}$ & $ 73.20  \pm 0.019 $ & $ 74.11 \pm 0.018 $ & $ 73.64 \pm 0.011 $ & $ 56.57  \pm 0.061 $ & $ \mathbf{75.59  \pm 0.017 }$\\ \hline
        JK-Net & Original & Resampling & Layerwise & Dropedge  & Dropout 0.2 \\ \hline
        $2\text{-Layers}$ & $ 58.43 \pm 0.020 $ & $ \mathbf{70.88  \pm 0.010 }$ & $ 69.92 \pm 0.018 $ & $ 60.35 \pm 0.037 $ & $ 59.61 \pm 0.031 $\\ 
        $4\text{-Layers}$ & $ 59.64 \pm 0.043 $ & $ \mathbf{69.24 \pm 0.013} $ & $ 68.94 \pm 0.020 $ & $ 55.36  \pm 0.070 $ & $ 59.26 \pm 0.020 $\\ 
        $8\text{-Layers}$ & $ 57.49  \pm 0.037 $ & $ \mathbf{69.09  \pm 0.017} $ & $ 65.10  \pm 0.024 $ & $ 46.28  \pm 0.057 $ & $ 58.59  \pm 0.029 $\\ 
        $16\text{-Layers}$ & $ 57.43 \pm 0.034 $ & $ \mathbf{60.98 \pm 0.042} $ & $ 54.73 \pm 0.042 $ & $ 46.55 \pm 0.080 $ & $ 58.30  \pm 0.025 $\\
        \hline
    \end{tabular}
    \egroup
    \label{tab:public}
\end{table}

\section{Conclusion} \label{sec:conc}
In this work, we introduced a novel and efficient graphon estimation technique for training deep Graph Neural Networks. Our proposed method augments the input graph to alleviate over-fitting and over-smoothing by drawing random adjacency matrix from the estimated graphon. Considerable experiments  on Cora, Citeseer and Pubmed on different splits have agreed that our graphon estimation method is able to promote the performance of several popular GNNs, like GCN, JKNet and GraphSAGE, in particular for the network with deep layers. To the best of our knowledge, this is  the first work utilizing graphon estimation on Graph Neural Networks. We also aim to exploit the theoretical analysis and large scale graph training of GNNs, with a variety of  graphon estimation methods in the future work.

\section*{Acknowledgment}
This research is partially supported by NSF grants DMS Career 1654579,  DMS 1854779 and DMS 2113642.

\printbibliography

\end{document}